\documentclass[letterpaper, 10 pt, journal, twoside]{IEEEtran} 
\IEEEoverridecommandlockouts                              

\usepackage{graphicx} 
\usepackage[table, xcdraw, usenames, dvipsnames]{xcolor}
\usepackage{subfiles}
\usepackage{soul}
\usepackage{multirow}
\usepackage{amsmath}
\usepackage{amssymb}
\usepackage{bm}
\usepackage{csquotes}

\title{Deep Modeling of Non-Gaussian \\Aleatoric Uncertainty}

\author{Aastha Acharya$^{1, 2, 3}$, Caleb Lee$^1$, Marissa D'Alonzo$^1$, Jared Shamwell$^1$, Nisar R. Ahmed$^2$, and Rebecca Russell$^1$
\thanks{Manuscript received: June 26, 2024; Revised October 9, 2024; Accepted November 8, 2024.}
\thanks{This paper was recommended for publication by Editor Aleksandra Faust upon evaluation of the Associate Editor and Reviewers' comments.
This research was, in part, funded by the U.S. Government. The views, opinions, and/or findings expressed are those of the author(s) and should not be interpreted as representing the official views or policies of the Department of Defense or the U.S. Government.}
\thanks{$^{1}$A.~Acharya, C.~Lee, M.~D'Alonzo, J.~Shamwell, and R.~Russell are with the Charles Stark Draper Laboratory, Inc., Cambridge, MA.}
\thanks{$^{2}$A.~Acharya and N.~Ahmed are with the Ann and H.~J.~Smead Department of Aerospace Engineering Sciences at the University of Colorado, Boulder, Boulder, CO.}
\thanks{$^{3}$A.~Acharya is a Draper Scholar funded by Draper.}
\thanks{Digital Object Identifier (DOI): see top of this page.}
}

\begin{document}

\markboth{IEEE Robotics and Automation Letters. Preprint Version. Accepted November, 2024}
{Acharya \MakeLowercase{\textit{et al.}}: Deep Modeling of Non-Gaussian Aleatoric Uncertainty} 

\maketitle


\begin{abstract}

Deep learning offers promising new ways to accurately model aleatoric uncertainty in robotic state estimation systems, particularly when the uncertainty distributions do not conform to traditional assumptions of being fixed and Gaussian.
In this study, we formulate and evaluate three fundamental deep learning approaches for conditional probability density modeling to quantify non-Gaussian aleatoric uncertainty: parametric, discretized, and generative modeling.
We systematically compare the respective strengths and weaknesses of these three methods on simulated non-Gaussian densities as well as on real-world terrain-relative navigation data.
Our results show that these deep learning methods can accurately capture complex uncertainty patterns, highlighting their potential for improving the reliability and robustness of estimation systems.

\end{abstract}

\section{INTRODUCTION}

Uncertainty is an inherent part of robots operating in a complex, partially-observable, and dynamic world. 
Probabilistic approaches to robotics that account for uncertainties in sensing, actuation, and the environment have led to significant advancements in the field~\cite{Thrun_2000}.
Our focus is on \emph{aleatoric uncertainty}, irreducible uncertainty due to sensor noise, stochasticity in the environment, lack of observability, or inherent ambiguity~\cite{alea_or_epist}.
This uncertainty can take complicated non-Gaussian forms and is typically \emph{heteroscedastic} (not constant) in real-world systems, particularly those operating with highly non-linear measurements/dynamics, high-dimensional sensor modalities, or incorporating deep learning components.

\begin{figure}[t]
    \centering
    \includegraphics[width=0.95\columnwidth]{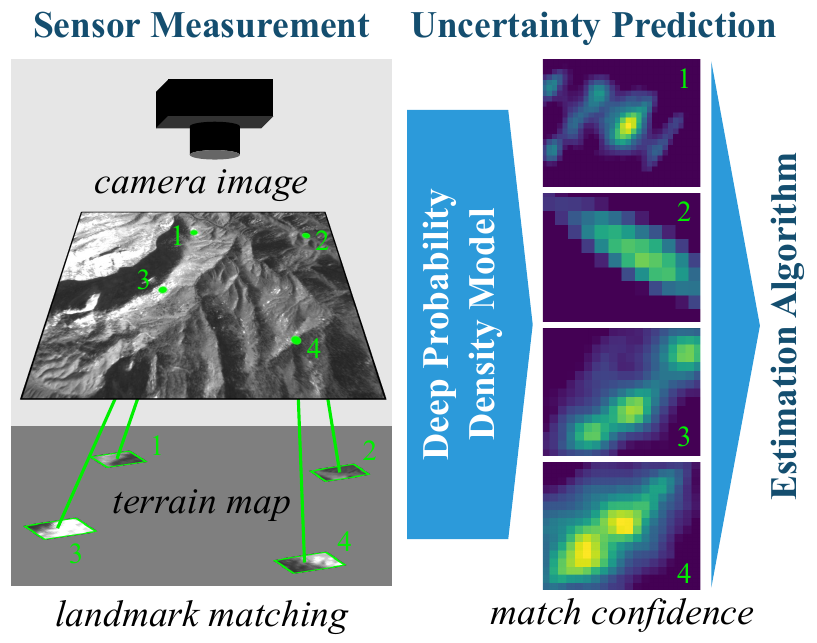}
    \caption{Illustration of deep uncertainty prediction for landmark matches in a visual terrain-relative navigation application. The predicted non-Gaussian probability densities can allow the estimation algorithm to better fuse uncertain information.
    \vspace*{-\baselineskip}}
    \label{fig:summary}
\end{figure}

Accurate aleatoric uncertainty modeling is important for sensor fusion and state estimation in robotics and usually requires characterization of measurement and process uncertainties directly from real world data.
The form of the probability density function (PDF) describing the  aleatoric uncertainty at a given time is often assumed to be Gaussian, uncorrelated, and constant, as is the case in a typical Kalman Filter \cite{kalman_1960}.
However, these assumptions are insufficient and potentially catastrophic when the PDFs are heteroscedastic, correlated, multi-modal, asymmetric, or have high kurtosis due to \enquote{long tail} events, as encountered in the real world \cite{nongausskalman}. 
For example, in the terrain relative navigation application illustrated in Figure~\ref{fig:summary}, the four landmark match measurements shown exhibit highly heteroscedastic and non-Gaussian PDFs.
As is typical in robotics applications, multimodal and high kurtosis PDFs appear due to multiple plausible solutions and irreducible ambiguity given the limited observable data.

Deep learning is state-of-the-art for density estimation with large training datasets~\cite{Papamakarios2019NeuralDE} but is not well-explored for state estimation and sensor fusion systems.
In this work, we analyze, compare, and evaluate three fundamental deep learning  approaches for accurately quantifying non-Gaussian heteroscedastic aleatoric uncertainty in a state estimation system.
Our models are trained directly on system residual data (given true state or measurement information) and approximate the PDFs of those residuals conditioned on the current system context (e.g., the raw sensor data), enabling them to capture heteroscedastic uncertainties.
We make minimal assumptions regarding the underlying PDFs and experiment on varying dimensions with different formats of conditional inputs, ensuring our results are generalizable to a large variety of robotics applications.

Accurate modeling of heteroscedastic non-Gaussian uncertainties enables summarization of the system uncertainty characteristics and quantification of the impact of any simplifying assumptions (e.g., Gaussian model) used in the system algorithms.
This uncertainty modeling also allows system designers to leverage the full potential of more advanced probabilistic algorithms~\cite{gordon1993, nongaussianfactor} such as  particle filters that can directly use a free-form measurement PDF.
Accurate uncertainty modeling will allow the creators of robotic systems to account for and be robust to these non-Gaussian uncertainty characteristics.

We contribute: (i) a formulation and demonstration of three fundamental approaches to non-Gaussian error modelling using deep learning on a real terrain-relative navigation application, (ii) an analytic error distribution simulation with adjustable dimensionality, modality, skew, kurtosis, and condition format to systematically evaluate these three uncertainty modeling approaches, and (iii) discussion of strengths and weaknesses of the three density estimation methods, including their suitability to different problems in state estimation. 
The three deep learning approaches we evaluate and compare for aleatoric uncertainty quantification are \emph{parametric}, \emph{discretized}, and \emph{generative}.

\section{BACKGROUND AND RELATED WORK} \label{relworks}

Density estimation is the process of modeling the PDF $p(\bm{\epsilon})$ given samples of the random variable $\bm{\epsilon} \in \mathbb{R}^K$ drawn from that distribution. 
Density estimation can be used in robotics for modeling noise or aleatoric uncertainty from data.
In conditional density estimation, we model $p(\bm{\epsilon}|\bm{x})$, the PDF conditioned on variable $\bm{x} \in X$, allowing the modeled distributions to be heteroscedastic.
In this work, we let $\bm{\epsilon}$ represent the $K$-dimensional residual between the true and the predicted (or measured) value, so $p(\bm{\epsilon}|\bm{x})$ represents the uncertainty of the measurement or prediction in the system.
Then, the conditional variable $\bm{x}$ is the observable input that correlates with the uncertainty distribution. 

While uncertainty quantification in robotics typically assumes Gaussian distributions~\cite{slam_algs},
non-Gaussian density estimation is performed using kernel density estimation (KDE) for applications such as path planning \cite{decentralized_kde} and localization \cite{kde_localization}.
$\epsilon$-neighbor KDE and its variations \cite{eKDE1, eKDE2} can perform conditional density estimation but require locally estimating the full joint distribution $p(\bm{\epsilon},\bm{x})$ and do not work well on high-dimensional data.  
Other kernel methods such as least squares conditional density estimation (LS-CDE)~\cite{lscde} scale better but are not as expressive due to the simplifying assumptions they must make. 
Recently, data-driven methods such as deep learning methods are being used for conditional density estimation due to both their expressibility and scalability to high dimensions~\cite{nn_cde}.

In this work, we formulate best practices and compare various existing deep learning methods of performing conditional density estimation in their current forms in the context of robotic state estimation applications.
While other works have evaluated epistemic (model-specific) uncertainty quantification methods~\cite{Arnez2020ACO}, non-Gaussian aleatoric uncertainty has been mostly overlooked despite its significance in big data applications.
We study three general deep learning approaches that have broad applicability to aleatoric uncertainty modeling in robotics: parametric, discretized, and generative.

The parametric Gaussian approach, wherein a neural network outputs the parameters of a Gaussian distribution approximating the target distribution, is the simplest deep learning method to model aleatoric uncertainty.
After its demonstrated success in capturing univariate aleatoric uncertainty for a variety of applications \cite{kendall_gal}, it was adapted to the multivariate case to enable its usage for tracking and navigation \cite{russell2021multivariate}.
To model non-Gaussian distributions, the neural network outputs the parameters of a more expressive parametric model, such as a Gaussian mixture model~\cite{Bishop1994MixtureDN}.
Our work analyzes a multivariate extension of this Gaussian mixture model formulation.

The discretized approach to performing deep conditional density estimation discretizes the continuous residual space $\mathbb{R}^K$, reframing it as a categorical distribution modeling problem~\cite{scott1992multivariate}.
Variations of this method include deconvolutional density networks~\cite{deconv_density_net}, alternative cross-entropy loss formulations \cite{deep_dist_regression}, or hierarchical approaches \cite{tansey2016better}.
Our work analyzes a generalized form of these methods~\cite{deconv_density_net, deep_dist_regression}. 

Finally, deep generative modeling, using either approximate methods such as variational autoencoders (VAEs)~\cite{Kingma2014} or exact methods such as normalizing flow (NF)~\cite{realnvp}, is a powerful approach to density estimation that is uniquely suited to high-dimensional and highly non-Gaussian distributions.
Examples of NFs are masked autoregressive flow \cite{maf}, inverse autoregressive flow \cite{iaf}, and real-valued non-volume preserving (RealNVP) model \cite{realnvp}.
This work analyzes RealNVP \cite{realnvp} due to its ability to capture exact distributions and its efficiency in performing parallel sampling.

These three deep learning approaches---parametric, discretized and generative---have been independently studied for the general problem of density estimation. However, there is a lack of understanding as to which are best suited to various sensor fusion and state estimation applications and how they should be appropriately implemented. This work seeks to fill that gap.

\section{METHODS}\label{methods}

In this section, we formalize and present best practices for the parametric, discretized, and generative approaches to using deep learning to model conditional probability densities.
All deep PDF models are conditioned on an input $\bm{x} \in X$ and model the distribution $p(\bm{\epsilon}|\bm{x})$ of the $K$-dimensional residuals $\bm{\epsilon} \in \mathbb{R}^K$.
The uncertainty ``condition'' $\bm{x}$ can be made up of any observable features that correlate with the aleatoric uncertainty of the system, e.g., input sensor data in the form of a vector or image.
All three classes of uncertainty model are trained through maximum likelihood estimation of the training dataset $\{(\bm{x}, \bm{\epsilon})\}$.
These uncertainty models all allow us to calculate the likelihood $p(\bm{\epsilon}|\bm{x})$ and efficiently draw samples $\bm{\epsilon} \sim p(\bm{\epsilon}|\bm{x})$, both of which may be useful in estimation applications.

\subsection{Parametric: Gaussian Mixture Density Model}
In the parametric uncertainty modeling approach, a deep neural network predicts $M$ parameter values $\bm{\alpha} \in \mathbb{R}^M$ of a parameterized probability distribution $p_\alpha(\bm{\epsilon}|\bm{x})$ from the input condition $\bm{x}$.
We start by assuming $p_\alpha(\bm{\epsilon}|\bm{x})$ is a multivariate Gaussian distribution.
As in Ref.~\cite{russell2021multivariate}, the neural network has two output heads corresponding to the distribution mean $\bm{\mu} : X \to \mathbb{R}^K$ prediction and covariance $\bm{\Sigma} : X \to \mathbb{R}^{K \times K}$ prediction, where $\bm{\Sigma}(\bm{x})$ must be symmetric $\bm{\Sigma}(\bm{x}) = \bm{\Sigma}(\bm{x})^\mathsf{T}$ and positive semi-definite $\bm{\Sigma}(\bm{x}) \succcurlyeq 0$.
Our implementation consists of an encoder made up of alternating 2D convolution, batch normalization, and ReLU layers that compresses the multidimensional input into the desired output size. This is followed by several linear layers and the post-processing that formats the raw neural network outputs into the PDF parameters.

The model is trained by minimizing the negative log likelihood of the observed residual data given the predicted PDF:
\begin{align}
L_G(\bm{x}, \bm{\epsilon}) & = \frac{1}{2}\left(\bm{\epsilon}-\bm{\mu}(\bm{x})\right)^\mathsf{T}\bm{\Sigma}(\bm{x})^{-1}\left(\bm{\epsilon}-\bm{\mu}(\bm{x})\right) \nonumber \\
& \hphantom{{}=} + \frac{1}{2}\ln\left|\bm{\Sigma}(\bm{x})\right|
+ \frac{K}{2}\ln\left(2\pi\right)
\label{eq:loss_func}
\end{align}

To enforce the constraints on the predicted covariance matrix, the network predicts the entries of the real lower unit triangular matrix $\mathbf{L}$ and of the positive diagonal matrix $\mathbf{D}$ that make up the LDL decomposition\footnote{We found that using this LDL decomposition of the covariance enabled faster and more stable learning than using a Cholesky decomposition or a Pearson correlation coefficient parameterization.} of the covariance $\bm{\Sigma} = \mathbf{L} \mathbf{D} \mathbf{L}^\mathsf{T}$.
We apply an exponential activation to the model outputs for the diagonal entries of $\mathbf{D}$ to enforce positivity.
The multivariate Gaussian distribution requires the neural network to predict a total of $M_G = K(K+3)/2$ independent parameter values $\bm{\alpha}$ to capture the means, variances, and correlations between the $K$ dimensions.

\begin{figure}
    \centering
    \includegraphics[width=0.95\columnwidth]{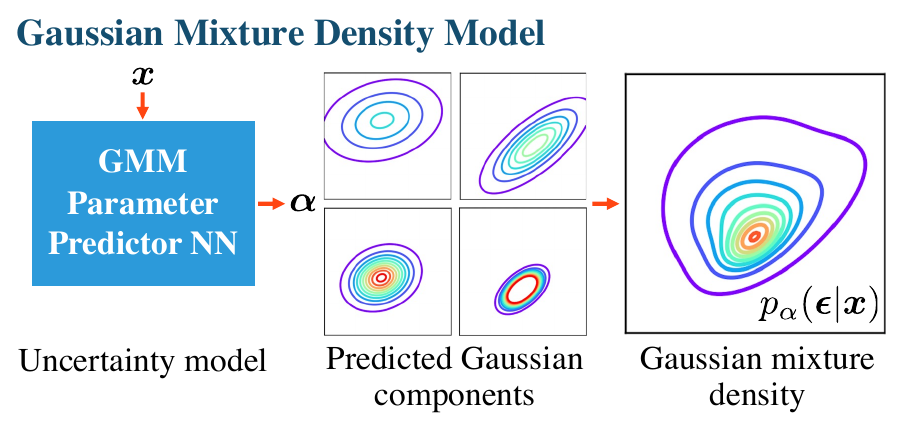}
    \caption{The parametric GMM uncertainty model uses a neural network (NN) to directly predict the parameters $\bm{\alpha}$ of components of a Gaussian mixture density $p_\alpha(\bm{\epsilon}|\bm{x})$, given input condition $\bm{x}$. 
    \vspace*{-\baselineskip}}
    \label{fig:gmm_overview}
\end{figure}

To model non-Gaussian multivariate probability densities, we parameterize $p_\alpha(\bm{\epsilon}|\bm{x})$ by a full Gaussian mixture model (GMM), illustrated in Figure~\ref{fig:gmm_overview}, a multivariate extension of Ref.~\cite{Bishop1994MixtureDN}.
The neural network then outputs parameters representing a mixture of $N$ Gaussian distributions with $N$ corresponding weights, for a total of $M_{\textit{GMM}} = N(M_G + 1)$ parameter values $\bm{\alpha}$.
We apply a softmax activation to get the parameters $w_n(\bm{x})$ representing the predicted weights of the individual Gaussian modes.
The GMM negative log-likelihood loss is then given in terms of the $n$th Gaussian component loss $L_G^n(\bm{x}, \bm{\epsilon})$ by:
\begin{align}
L_\textit{GMM}(\bm{x}, \bm{\epsilon}) = -\ln \left[ \sum_{n=1}^N w_n(\bm{x}) e^{-L^n_G(\bm{x}, \bm{\epsilon})}\right]
\label{eq:gmm_loss}
\end{align}
Optionally, we can also add a predicted background probability $w_{N+1}(\bm{x})$ to the mixture density for applications with a restricted $\bm{\epsilon}$ domain.
This background probability is particularly helpful for highly sparse probability densities, as in our terrain-relative navigation application (Section~\ref{trn}).

The GMM can approximate any smooth density given a high enough $N$, which we leave as a hyperparameter to be tuned. We expect this uncertainty model to be particularly accurate when modeling multi-modal distributions, in any number of dimensions.

\subsection{Discretized Density Model}

\begin{figure}[t]
    \centering
    \includegraphics[width=0.95\columnwidth]{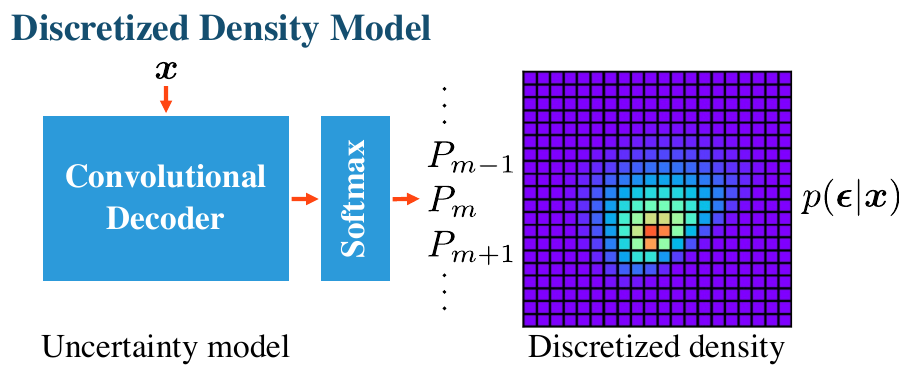}
    \caption{The discretized uncertainty model uses a decoder-style (upsampling) convolutional neural network to predict the probabilities $P_m$ of bins in a histogram representing the conditional probability density $p_\alpha(\bm{\epsilon}|\bm{x})$.
    \vspace*{-\baselineskip}}
    \label{fig:dcm_overview}
\end{figure}

The discretized uncertainty modeling approach, illustrated in Figure~\ref{fig:dcm_overview}, models continuous probability densities as categorical distributions over a discretized, or binned, version of the space.
Generalizing the approaches in Refs.~\cite{deconv_density_net, deep_dist_regression}, our model predicts the probability of an observed residual falling into each density ``histogram'' bin, which is equivalent to a multi-class classification problem with the classes corresponding to $M$ bins.

We must define the mapping $m(\bm{\epsilon})$ of our samples $\bm{\epsilon} \in \mathbb{R}^K$ to the bin indices. For each dimension $k$, we define the upper bound $b^+_k \in \mathbb{R} \mid b^+_k > b^-_k$, the lower bound $b^-_k \in \mathbb{R}$, and the number of bins per dimension $N_k \in \mathbb{P}$.
The training dataset is restricted to $\epsilon_k \in (b_k^-, b_k^+)$.
The neural network with input $\bm{x}$ must have $M = \prod_{k=1}^K N_k$ outputs, which we apply a softmax activation over to obtain the predicted categorical bin probabilities $P_m(\bm{x})$ where $\sum_{m=1}^M P_m(\bm{x}) = 1$.

The bin index $n_k \in \{1, \ldots, N_k\}$ along dimension $k$ that an observed residual $\epsilon_k$ falls into is then given by: 
\begin{equation}
n_k(\epsilon_k) = \left\lceil\left(\frac{\epsilon_k - b^-_k}{b^+_k - b^-_k}\right) N_k \right\rceil,
\end{equation}
and the overall bin index $m \in \{1, \ldots, M\}$ that the observed residual vector $\bm{\epsilon}$ falls into is given by: 
\begin{equation}
m(\bm{\epsilon}) = 1 + \sum_{k=1}^K \left(\prod_{j=k+1}^K \! N_j\right) \left(n_k(\epsilon_k) - 1\right).
\end{equation}
Note that the number of bins $M$ increases exponentially with the number of dimensions $K$ for a fixed binning density.

As in standard multi-class classification, the loss function given by the negative log likelihood of the distribution is the cross-entropy (negative log-likelihood) loss for the true bin:
\begin{equation}
    L_d(\bm{x}, \bm{\epsilon}) = -\ln{P_{m(\boldsymbol{\epsilon})}(\bm{x})}
\end{equation}
The predicted probabilities for each bin form a density histogram as shown in Figure \ref{fig:dcm_overview}.

In many applications, particularly those using computer vision which care about measurements down to the pixel, we need a high binning density to achieve the uncertainty quantification precision required.
To increase data efficiency, we use a decoder-style convolutional neural network architecture that mixes convolutional layers with upsampling operations to go from an encoded vector to a high-resolution probability density map. 
This convolutional decoder architecture adds an important inductive bias to the model to predict smoother densities.

The discretized uncertainty models quickly converge to smooth probability density histogram predictions during training and generalize well to unseen condition inputs, but are only practical for $K \leq 3$ due to the curse of dimensionality and challenge in implementing higher dimensional convolutions.
Additionally, since discretized models are inherently bounded, they cannot predict deep into the tails of distributions as accurately as the other, unbounded models, when the measurement range is unbounded.

\subsection{Generative: Normalizing Flow Density Model}

The generative uncertainty modeling approach trains a model $f : \mathbb{R}^M \times X \to \mathbb{R}^K$ to map samples $\bm{z} \in \mathbb{R}^M$ from a known latent probability distribution $\pi(\bm{z})$ to the data generating distribution $p(\bm{\epsilon} | \bm{x})$.
Typically, the latent distribution $\pi(\bm{z})$ is a unit Gaussian distribution.
As in our previous two methods, the training of $f$ aims to maximize $\ln p(\bm{\epsilon}|\bm{x})$ of the training dataset $\{(\bm{x},\bm{\epsilon})\}$, though calculating this criterion for generative models is less straightforward than it is for the discriminative methods.
Generative models allow us to capture and sample from high-dimensional probability distributions without a parametric representation of the density.

\begin{figure}[t]
    \centering
    \includegraphics[width=0.95\columnwidth]{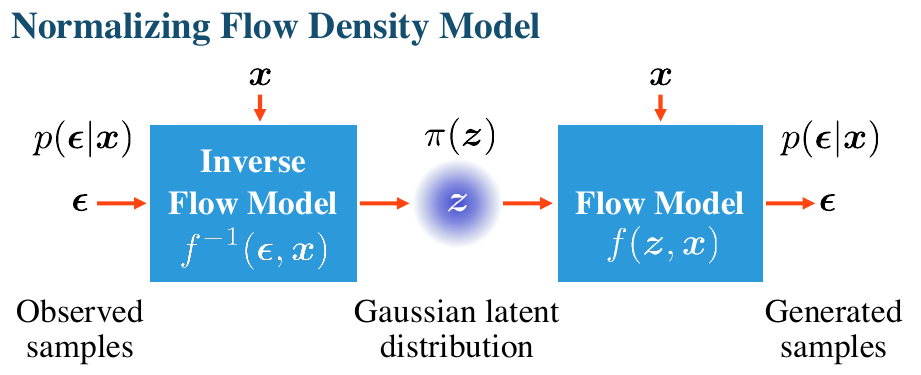}
    \caption{The conditional normalizing flow density modeling approach uses the parameterized invertible variable transformations that make up a flow model, conditioned on $\bm{x}$, to map a latent distribution $\pi(\bm{z})$ to the data generating distribution $p(\bm{\epsilon}|\bm{x})$.
    \vspace*{-\baselineskip}}
    \label{fig:nflow_overview}
\end{figure}

While conditional variational autoencoders (VAEs) can be used to model aleatoric uncertainty in high-dimensional distributions~\cite{acharya2022competency, acharya2023learning}, they cannot calculate an exact likelihood $p(\bm{\epsilon} | \bm{x})$ and are challenging to train in lower dimensions or under strong conditioning.
Instead, we use conditional normalizing flow (NF) models~\cite{papamakarios2021normalizing} for precise density  modeling, as shown in Figure~\ref{fig:nflow_overview}.
Conditional NFs transform the latent distribution $\pi(\bm{z})$ into the data distribution $p(\bm{\epsilon}|\bm{x})$ by applying a sequence of invertible, parameterized transformation functions that are themselves conditioned on $\bm{x}$.
The learned mapping $\bm{\epsilon} = f(\bm{z}, \bm{x})$ allows us to generate samples from $p(\bm{\epsilon}|\bm{x})$ by transforming samples $\bm{z} \sim \pi(\bm{z})$ given $\bm{x}$.

Since each layer of $f$ is designed to be invertible, $f$ as a whole is invertible with respect to $\bm{\epsilon}$, so $\bm{z} = f^{-1}(\bm{\epsilon}, \bm{x}$).
The restriction of the transformation layers being invertible seriously limits the architectural choices for NFs and requires that the dimensionality of $\bm{z}$ equal the dimensionality of $\bm{\epsilon}$, $M=K$.
However, this restriction and the requirement to be able to easily calculate the Jacobian determinants of the layers also allow for an efficient and exact calculation of the likelihood of data from the inverse model and its Jacobian determinant using the change of variables theorem:
\begin{align}
    p(\bm{\epsilon} | \bm{x}) & = \pi(\bm{z}) \left|\det\left(\frac{d\bm{z}}{d\bm{\epsilon}}\right)\right| \nonumber \\
    & = \pi\left(f^{-1} (\bm{\epsilon}, \bm{x})\right)\left|\det\left(\frac{df^{-1}}{d\bm{\epsilon}}(\bm{\epsilon}, \bm{x})\right)\right|
    \label{eq:nf_likelihood}
\end{align}
The negative log likelihood loss function used to train the NF model is then simply:
\begin{align}
L_{\textit{NF}}(\bm{x}, \bm{\epsilon}) & = -\ln\pi\left(f^{-1} (\bm{\epsilon}, \bm{x})\right) \nonumber \\
& \hphantom{{}=} - \ln\left|\det\left(\frac{df^{-1}}{d\bm{\epsilon}}(\bm{\epsilon}, \bm{x})\right)\right|
\end{align}

In this work, we use the real-valued non-volume preserving (RealNVP) normalizing flow architecture~\cite{realnvp}, which is designed for density estimation. It is able to capture exact distributions while performing parallel sampling in an efficient manner.

\subsection{Metrics}

We evaluate two quantities, with complementary strengths and weaknesses, in our experiments to compare and assess the three approaches to uncertainty modeling:

\subsubsection{Negative Log-Likelihood (NLL)}
The average negative logarithm of the likelihood of the observed residuals given the predicted conditional probability densities.
This is the best quantity for comparison \emph{between} probabilistic models and it reflects all the uncertainty characteristics we want to capture \cite{mackay2003information}.
However, NLL values can range from $-\infty$ to $\infty$ in non-intuitive units, which makes them difficult to interpret in the standalone evaluation of a model.

\subsubsection{Hellinger Distance (H)}
An $f$-divergence between the density predicted by the uncertainty model and the true density, ranging from 0 to 1 \cite{beran1977minimum}.
To evaluate the Hellinger distance, we discretize the $K$-dimensional residual space and numerically calculate the average density in each bin, which is only practical to do for low-dimensional uncertainty problems. For two discrete distributions $P = (p_1, ..., p_k)$ and $Q = (q_1, ..., q_k)$, the Hellinger distance $H$ can be defined as:
\begin{equation}
    H(P, Q) = \frac{1}{\sqrt{2}}  \sum_{i=1}^{k} \sqrt{(\sqrt{p_{i}} - \sqrt{q_{i}})^2}
\end{equation}
The NLL can be evaluated exactly from data alone, without knowledge of the true underlying conditional densities.
When the true conditional densities are known, as in our simulation experiments, we average over the Hellinger distance values in the test set.
When the true conditional densities are unknown and we only have the observed residuals, as is the case in any real application, we calculate the Hellinger distance from the average of the conditional densities the observed distribution of the residuals accumulated over the test set.
This latter calculation can lose information about the model's ability to capture conditional information, which is why it is important to pair the Hellinger distance with the log-likelihood value.

\section{SIMULATION EXPERIMENTS} \label{simexp}

\newcommand{\mc}[3]{\multicolumn{#1}{#2}{#3}}
\newcommand{\nll}{\mc{1}{c}{NLL}}
\newcommand{\hel}{\mc{1}{c|}{$\bar{H}$}}
\renewcommand{\arraystretch}{1.2}
\begin{table*}[t]
\caption{\label{tab:sim_results}Evaluation results for simulation experiments}
\centering
\begin{tabular}{c||cc|cc|cc||cc|cc|cc}
\hline
& \mc{6}{c||}{\textbf{Multimodal scalar-conditioned experiments}} & \mc{6}{c}{\textbf{Parameter- and image-conditioned experiments}} \\
& \mc{2}{c|}{$K=1$} & \mc{2}{c|}{$K=2$} & \mc{2}{c||}{$K=6$} & \mc{2}{c|}{Param. ($K=1$)} & \mc{2}{c|}{Param. ($K=2$)} & \mc{2}{c}{Image ($K=2$)} \\
\textbf{Uncertainty Model} & \nll & \hel & \nll & \hel & \nll & \mc{1}{c||}{$\bar{H}$} & \nll & \hel & \nll & \hel & \nll & \mc{1}{c}{$\bar{H}$} \\ \hline
Parametric Gaussian & 0.688 & 0.116 & 1.432 & 0.254 & 3.247 & n/a & 0.376 & 0.149 & 0.809 & 0.128 & 0.956 & 0.133 \\
Parametric GMM      & \textbf{0.634} & \textbf{0.041} & 1.203 & 0.101 & 2.049 & n/a & \textbf{0.314} & \textbf{0.033} & \textbf{0.747} & 0.076 & \textbf{0.894} & 0.097 \\
Discretized         & 0.645 & 0.061 & \textbf{1.171} & \textbf{0.032} & n/a           & n/a & 0.338 & 0.057 & 0.946 & 0.079 & 1.245 & \textbf{0.096} \\
Generative NF       & 0.901 & 0.070 & 1.194 & 0.123 & \textbf{1.927} & n/a & 0.514 & 0.085 & 0.944 & \textbf{0.056} & 1.117 & 0.104 \\ \hline
\textbf{Perfect Reference} & 0.626 & 0 & 1.160 & 0 & 1.834 & 0 & 0.310 & 0 & 0.724 & 0 & 0.790 & 0 \\ \hline
\end{tabular}
\end{table*}

We are interested in evaluating the ability of deep learning uncertainty models to capture the following oft-neglected uncertainty characteristics in a distribution:
\begin{itemize}
    \item \textbf{Correlation:} The dependence of the distribution in one dimension on another (non-zero product moment).
    \item \textbf{Heteroscedasticity:} The variation of uncertainty dependent on some condition (conditional 2nd moment).
    \item \textbf{Skewness:} The asymmetry of the distribution (non-zero 3rd moment).
    \item \textbf{Kurtosis:} The heaviness of the uncertainty distributions tails  (non-zero 4th moment).
    \item \textbf{Multimodality:} The existence of more than one local maximum (peaks) in the distribution.
\end{itemize}
To compare the models against ground truth, we develop an error simulation with a closed-form probability density function that allows each of these characteristics to be varied. 

\subsection{Analytic Error Simulation}

Our error simulation is based on the Skewed Generalized Error Distribution (SGED), which is a generalization of a Gaussian distribution with variable skewness and kurtosis~\cite{sgtd}.
The SGED probability density function of an error residual $\epsilon$ is defined by: 
\begin{equation}
      f_{\text{SGED}}(\epsilon) =\frac{p}{2s \Gamma(1/p)}
    e^{-\left(\frac{|\epsilon|}{s\left(1 +
    \lambda\,\mathrm{sgn}(\epsilon)\right)}\right)^p},
\end{equation}
where $s > 0$ controls the scale, $-1 < \lambda < 1$ controls the skew, and $p > 0$ controls the kurtosis. The values $s=\sigma / \sqrt{2}$, $\lambda=0$, and $p=2$ recover the standard form of the Gaussian distribution, while $p=1$ corresponds to a Laplace distribution.

We also formulate a \emph{multivariate} version of the SGED, termed MV-SGED, to study the models' abilities to capture correlations between different dimensions and high-dimensional distributions. 
Our $K$-dimensional MV-SGED has a consistent kurtosis parameter $p$ across all dimensions but has independent scale parameters $s_k$ and skew parameters $\lambda_k$ across all (decorrelated) dimensions.
The uncorrelated-form multivariate SGED probability density function of our error residual $\epsilon$ and dimension \textit{n} is defined by:
\begin{align}
f_\text{MV-SGED}(\bm{\epsilon}) = & \frac{p \Gamma(K/2)}{2 \pi^{\frac{K}{2}} \prod_{i=1}^{K} s_i \Gamma(K/p)} \nonumber  \\
& \times e^{-\left[\sum_{k=1}^K\left(\frac{\epsilon_{k}}{s_k (1 + \lambda_k \mathrm{sgn}(\epsilon_k))}\right)^{2}\right]^{\frac{p}{K}}}
\end{align}
In order to add correlations between dimensions, we rotate the distribution over $\bm{\epsilon}$ by a rotation matrix from $\mathrm{SO}(K)$.
To simulate multimodality, we create a \textit{multimodal} version of the MV-SGED (MM-SGED), which is a distribution comprised of a normalized combination of $N$ randomly offset MV-SGEDs.

We evaluate our uncertainty models in various dimensions ($K=1, 2, 6$) with three forms of conditioning: 

\begin{itemize}
\item \textbf{Scalar-conditioned}: MM-SGEDs are independently sampled corresponding to different condition points $x \in (-1, +1)$. The distributions at conditions in between those points are generated by interpolating the MM-SGED parameters of the sampled condition points with Hermitian splines. Thus, the MM-SGED smoothly varies with the scalar condition $x$.

\item \textbf{Parameter-conditioned}: (Unimodal only) The uncertainty models are conditioned on a vector containing the true randomly-sampled MV-SGED parameters.

\item \textbf{Image-conditioned}: (2D only) The uncertainty models are conditioned on features extracted by a convolutional neural network from an image showing the true MV-SGED probability density.
\end{itemize} 

For each experiment, the training dataset consists of 100,000 randomly sampled conditions and, for each condition, a single randomly sampled point from the true distribution corresponding to that condition.
Thus, for all experiments, the uncertainty models must generalize across conditions to model the full conditional probability distributions.
All uncertainty models are trained with equivalent condition encoder architectures to ensure fair comparison.
Hyperparameters were chosen based on separate validation set.

\subsection{Simulation Experimental Results and Analysis}

Our full simulation evaluation results are shown in Table~\ref{tab:sim_results}, with NLL and average Hellinger distance evaluated over test sets of 1,000 unseen conditions, and the \enquote{perfect} results from evaluating the true distributions on the same data provided for reference.
Each model type excels most at representing different types of conditional probability densities.

In the scalar-conditioned results, we see that the parametric GMM, the discretized model, and the generative NF model each out-perform the rest in the 1-D, 2-D, and 6-D cases, respectively.
The GMM performed well in all dimensions but best in 1-D, where the fixed number of Gaussian modes had more relative flexibility to approximate the non-Gaussian shape of the true modes.
The discretized model performed best in the 2-D case, where its flexibility in approximating non-Gaussian peak shapes balanced best with its inability to handle higher dimensions---the latter being why it could not be evaluated in 6-D.
The generative NF uncertainty model performed best in 6-D and worst in 1-D, presumably due to the higher-dimensional flow architectures having increased flexibility from mixing dimensions.

In the parameter- and image-conditioned results, the GMM uncertainty model again shows strong across-the-board performance, most clearly excelling in the 1-D case.
The discretized model performed best on the image-conditioned experiment, likely because that condition encoded information about the distribution in a format similar to the discretized output.
While the generative NF model again performed poorly in the 1-D case, it performed reasonably well in the two 2-D cases despite only being able to combine two dimensions, likely in part to being able to better capture kurtosis than the other models.
Due to its flexibility, the generative NF model sometimes performed well in the Hellinger distance metric but not in NLL, even compared to the baseline---it did not have the strong inductive bias towards smooth PDFs that the other two approaches have.

\section{APPLICATION} \label{trn}

To study the accuracy and utility of deep aleatoric uncertainty modeling on real data in a practical application, we evaluate our three density modeling approaches on the problem of Terrain Relative Navigation (TRN).
TRN localizes a camera by matching visual terrain features from a camera to ``landmark'' satellite imagery features with known locations.
TRN has been broadly applied in the air and space domains, including on the US military's Joint Precision Airdrop System (JPADS) \cite{airdrop} and NASA's Safe and Precise Landing Integrated Capabilities Evolution (SPLICE) program \cite{splice}.

\subsection{TRN System}

\begin{figure}[t]
    \centering
    \includegraphics[width=0.9\columnwidth]{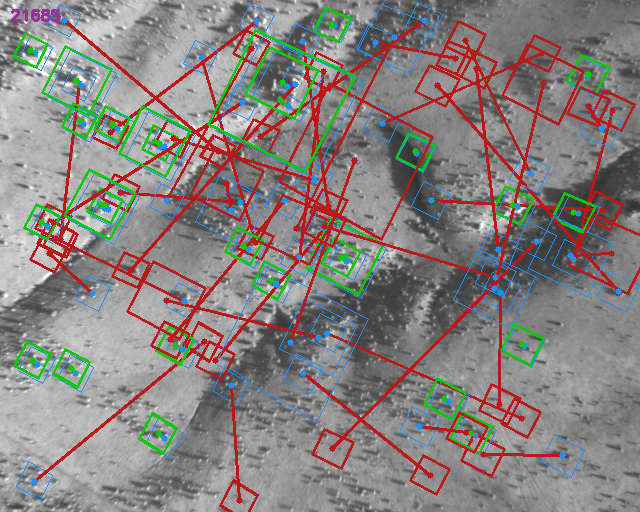}
    \caption{Inlier TRN match locations (green squares) and outlier TRN match locations (red squares) shown connected to their corresponding predicted locations (blue squares). The TRN system studied uses PnP RANSAC to reject bad matches, which can be overwhelmed when few good matches exist.
    \vspace*{-\baselineskip}}
    \label{fig:trn_example}
\end{figure}

TRN requires computing the visual similarity between satellite image landmark patches and camera image patches.
This similarity can be calculated with a classical template matching measure such as normalized cross-correlation (NCC) or with a deep similarity learning embedding.
Either approach to matching can have large uncertainties or completely fail when the camera image is different from the reference satellite imagery due to shadows, blur, occlusion, registration effects, or content change.
To increase robustness against these failures, our TRN system~\cite{maggio_ibal} uses many landmarks per frame when possible and checks the matches for consistency via perspective-n-point random sample consensus (PnP RANSAC) to reject outliers, as shown in Figure~\ref{fig:trn_example}.

Typically, TRN systems consider the uncertainty of each consistent match to be homogeneous, Gaussian, and uncorrelated, discarding high-uncertainty matches entirely.
Instead, we now apply our three deep uncertainty models to predict the match pixel uncertainty conditioned upon the similarity surface from the matching itself.
In this application, $\bm{\epsilon} \in \mathbb{R}^2$ is the 2D pixel residual of the similarity surface peak (match location) and the conditioning variable $\bm{x}$ is the similarity surface, a 2D one-channel image.
We chose to condition our uncertainty models on the similarity surfaces rather than the raw camera and landmark images to encourage increased generalization to new scenes and terrain.
Our experiments use a fully-convolutional neural network cosine distance metric learning model as the application system, though our methods are equally valid for non-learning template-based matching such as with NCC.

\subsection{TRN Evaluation Results}

We analyze our models on a TRN dataset from roughly five hours of flight at varying altitudes over Colorado containing diverse terrain types: urban, agricultural, grassland, forest, valley, lake, and mountain.
In total, the dataset contains roughly 300,000 pairs of 512x640 camera frames and variably-sized landmarks.
We determined ground truth for each landmark match location using a ``gold standard'' navigation solution from a navigation-grade GPS/inertial system aided by visual feature tracking.
The uncertainty models were trained and evaluated using 4-fold validation over the dataset.

\begin{table}[tb]
\renewcommand{\arraystretch}{1.2}
\caption{\label{tab:trn_results}Evaluation results for TRN experiments}
\centering
\begin{tabular}{c||c|c}
\hline
\textbf{Uncertainty Model} & NLL & H \\ \hline
Fixed Gaussian (benchmark) & 5.22 & 0.201 \\ 
Parametric GMM & 4.56 & 0.092 \\
Discretized & 3.24 & \textbf{0.044} \\
Generative NF & \textbf{2.99} & 0.088 \\ \hline
\end{tabular}
\end{table}

The evaluation results of our three uncertainty model types, compared to a fixed Gaussian distribution plus background around the similarity peak (current practice), are shown in Table~\ref{tab:trn_results}.
The parametric GMM model, which was trained with a variable constant background term and 3 modes, is able to capture the multimodality and 2D pixel correlations in the uncertainty, but struggles with the high kurtosis of the probability density peaks. 
The discretized and generative uncertainty models, on the other hand, are able to capture highly non-Gaussian peak shapes.
The generative uncertainty model likely achieves a better NLL than the discretized model because it models subpixel density, while the discretized model was binned at the pixel level.
Note that the TRN data provides some major additional challenges compared to the simulation experiments that explain the differences: the modes are very sparse and have extremely high kurtosis.

\begin{figure}[tb]
    \centering
    \includegraphics[width=0.85\columnwidth]{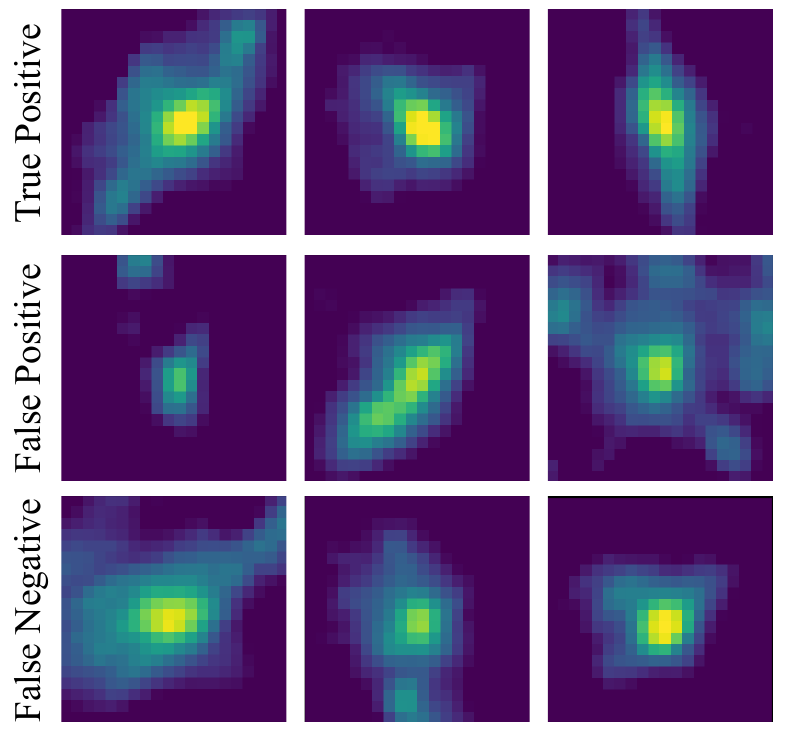}
    \caption{Sample probability density predictions (log scale) from the discretized uncertainty model around the locations of true positive matches (top row), false positive matches (middle row), and false negative or missed match locations (bottom row).
    \vspace*{-\baselineskip}}
    \label{fig:trn_prob_examples}
\end{figure}

Random examples of discretized model log-scale probability density predictions (20x20 pixel regions of interest within the full 512x650 prediction) are shown in Figure~\ref{fig:trn_prob_examples}. 
True/false positive/negative examples were determined based on whether the similarity peak was near the true landmark location.
The predicted true positive and false negative distributions show a more distinct probability peak than the false positive distributions, indicating that the uncertainty model can provide valuable information to the navigation system.
Observe that the discretized uncertainty model is able to capture multimodal and highly non-Gaussian distributions.
The parametric GMM uncertainty model and the NF uncertainty model are similarly able to capture the multimodality and peak non-Gaussianity, respectively.

\begin{figure}[tb]
    \centering
    \includegraphics[width=0.98\columnwidth]{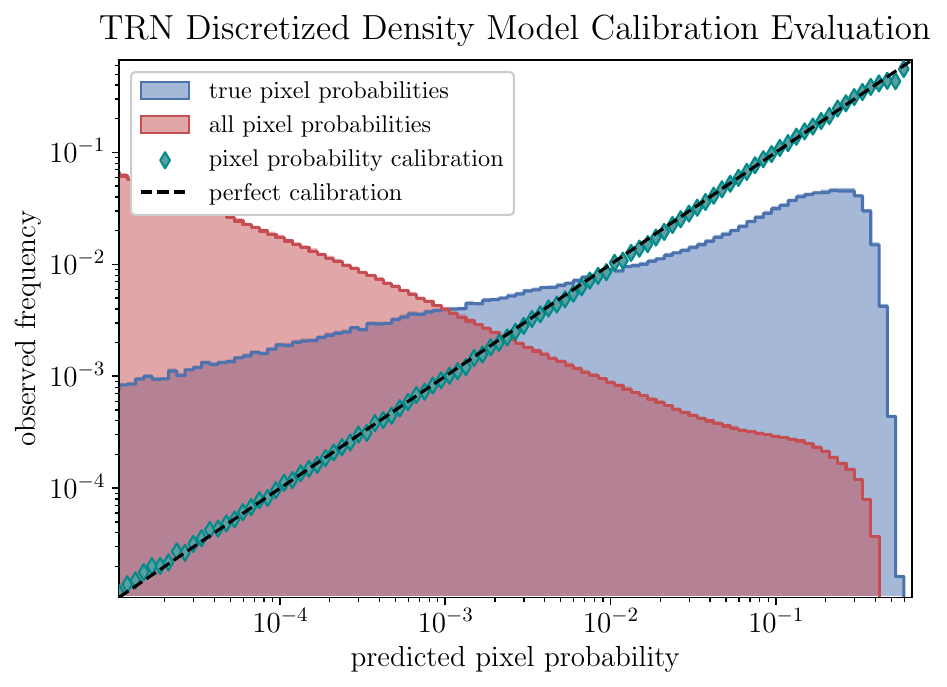}
    \caption{The observed true location frequency versus predicted pixel probability (circular cyan markers) from the discretized uncertainty model fall along the diagonal representing perfect probability calibration from $10^{-5}$ to $1$. The distribution of true pixel probabilities (light blue histogram) and all pixel probabilities (light red histogram) in the dataset are shown for reference. \vspace*{-\baselineskip}}
    \label{fig:trn_prob_calib}
\end{figure}

Finally, Figure~\ref{fig:trn_prob_calib} shows that the predicted pixel probabilities from the discretized model are well-calibrated over five orders of magnitude, indicating that low-probability predictions are just as accurate as high-probability predictions.
These results provide strong support that the deep learning uncertainty model predictions can be trusted to be accurate and used directly in a probabilistic estimation system.

Although the discretized model performed best in our experiments, other factors were considered when determining which model should be integrated with our existing TRN system.
All models performed better on our metrics than the benchmark, indicating that any choice would improve overall TRN performance.
The generative flow model was dismissed due to slower-than-realtime inference from the need to sample from the PDF.
The parametric GMM was chosen over the discretized model due to faster inference and because the covariance and weight parameters output could be directly integrated with the existing TRN system.
When running in-the-loop on our Colorado flight datasets, the parametric GMM reduced median frame distance error by 47\%.

\section{CONCLUSIONS} \label{conclusions}

We formulated and investigated three fundamental approaches to the deep modeling of non-Gaussian aleatoric uncertainty for robotics estimation problems: parametric, discretized, and generative.
We evaluated these approaches using an analytic conditional error simulation based on multivariate and multimodal generalizations of the Skew Generalized Error Distribution to understand the approaches' relative strengths and weaknesses in capturing various non-Gaussian uncertainty characteristics.
We additionally demonstrated the applicability of all three approaches to uncertainty quantification for a real-world visual terrain-relative navigation system and dataset.

Our formulated models and top-level conclusions based on these three approaches are:
\begin{enumerate}
\item \emph{Parametric Gaussian Mixture Density Model}: The neural network predicts the parameters of the Gaussian mixture density describing the aleatoric uncertainty. This approach is easy to train, works well in both high and low dimensions, and excels for multi-modal uncertainty distributions. It also tends to have the fastest inference times and be the easiest to integrate in estimation systems. It is weakest for high-kurtosis distributions.
\item \emph{Discretized Density Model}: The neural network predicts the probability of the true residual falling within each bin of a discretized output space. This approach only works for lower-dimensional spaces (one to three), but is highly accurate and particularly well-suited to applications that are naturally discretized.
\item \emph{Generative Normalizing Flow Density Model}: The neural network learns a conditional mapping between a $K$-dimensional Gaussian and the $K$-dimensional distribution defined by the data. This approach is highly accurate in high dimensions but is more challenging to train or to integrate into some existing systems.
\end{enumerate}
We have shown that all three approaches can be powerful tools for the modeling of non-Gaussian aleatoric uncertainty in estimation systems, with different approaches being more appropriate for different applications.
Future work should quantify the improvements in system reliability and robustness when non-Gaussian uncertainty quantification is accurately captured in a robotic system.

\addtolength{\textheight}{-3cm}   




\section*{ACKNOWLEDGMENT}
This material is based upon work supported by the Defense Advanced Research Projects Agency (DARPA) under Agreement No. HR0011-22-9-0114. A. Acharya's work was funded by Draper under the Draper Scholars program.

\bibliographystyle{IEEEtran}
\bibliography{IEEEabrv, references}

\begin{thebibliography}{10}
\providecommand{\url}[1]{#1}
\csname url@rmstyle\endcsname
\providecommand{\newblock}{\relax}
\providecommand{\bibinfo}[2]{#2}
\providecommand\BIBentrySTDinterwordspacing{\spaceskip=0pt\relax}
\providecommand\BIBentryALTinterwordstretchfactor{4}
\providecommand\BIBentryALTinterwordspacing{\spaceskip=\fontdimen2\font plus
\BIBentryALTinterwordstretchfactor\fontdimen3\font minus
  \fontdimen4\font\relax}
\providecommand\BIBforeignlanguage[2]{{%
\expandafter\ifx\csname l@#1\endcsname\relax
\typeout{** WARNING: IEEEtran.bst: No hyphenation pattern has been}%
\typeout{** loaded for the language `#1'. Using the pattern for}%
\typeout{** the default language instead.}%
\else
\language=\csname l@#1\endcsname
\fi
#2}}

\bibitem{Thrun_2000}
S.~Thrun, ``Probabilistic algorithms in robotics,'' \emph{AI Magazine},
  vol.~21, no.~4, p.~93, Dec. 2000.

\bibitem{alea_or_epist}
A.~D. Kiureghian and O.~Ditlevsen, ``Aleatory or epistemic? {D}oes it matter?''
  \emph{Structural Safety}, vol.~31, no.~2, pp. 105--112, 2009.

\bibitem{kalman_1960}
R.~E. Kalman, ``A new approach to linear filtering and prediction problems,''
  \emph{J. Basic Eng.}, vol.~82, no.~1, pp. 35--45, 1960.

\bibitem{nongausskalman}
M.~Raitoharju, R.~Piché, and H.~Nurminen, ``A systematic approach for
  {K}alman-type filtering with non-{G}aussian noises,'' in \emph{19th Int.
  Conf. Inf. Fusion (FUSION)}, 2016, pp. 1853--1858.

\bibitem{Papamakarios2019NeuralDE}
G.~Papamakarios, ``{Neural Density Estimation and Likelihood-free Inference},''
  Ph.D. dissertation, Edinburgh U., 2019.

\bibitem{gordon1993}
N.~J. Gordon, D.~J. Salmond, and A.~F. Smith, ``Novel approach to
  nonlinear/non-{G}aussian {B}ayesian state estimation,'' in \emph{IEE Proc. F
  (Radar and Signal Processing)}, vol. 140, no.~2, 1993, pp. 107--113.

\bibitem{nongaussianfactor}
D.~M. Rosen, M.~Kaess, and J.~J. Leonard, ``Robust incremental online inference
  over sparse factor graphs: Beyond the {G}aussian case,'' in \emph{IEEE Int.
  Conf. Robot. and Automat. (ICRA)}, 2013, pp. 1025--1032.

\bibitem{slam_algs}
H.~Durrant-Whyte and T.~Bailey, ``himultaneous localization and mapping: part
  {I},'' \emph{IEEE Robot. \& Automat. Mag.}, vol.~13, no.~2, pp. 99--110,
  2006.

\bibitem{decentralized_kde}
G.~Foderaro, S.~Ferrari, and M.~M. Zavlanos, ``A decentralized kernel density
  estimation approach to distributed robot path planning,'' in \emph{Proc.
  Neural Inf. Process. Syst. Conf. (NeurIPS)}, 2012.

\bibitem{kde_localization}
B.~Kr{\"o}se, N.~Vlassis, R.~Bunschoten, and Y.~Motomura, ``A probabilistic
  model for appearance-based robot localization,'' \emph{Image and Vis.
  Comput.}, vol.~19, no.~6, pp. 381--391, 2001.

\bibitem{eKDE1}
P.~Hall, R.~C.~L. Wolff, and Q.~Yao, ``Methods for estimating a conditional
  distribution function,'' \emph{J. American Statistical Assoc.}, vol.~94, no.
  445, pp. 154--163, 1999.

\bibitem{eKDE2}
J.~Fan, Q.~Yao, and H.~Tong, ``Estimation of conditional densities and
  sensitivity measures in nonlinear dynamical systems,'' \emph{Biometrika},
  vol.~83, no.~1, pp. 189--206, 1996.

\bibitem{lscde}
M.~Sugiyama, I.~Takeuchi, T.~Suzuki, T.~Kanamori, H.~Hachiya, and D.~Okanohara,
  ``Conditional density estimation via least-squares density ratio
  estimation,'' in \emph{13th Int. Conf. Artif. Intell. and Statist.}, 2010,
  pp. 781--788.

\bibitem{nn_cde}
J.~Rothfuss, F.~Ferreira, S.~Walther, and M.~Ulrich, ``Conditional density
  estimation with neural networks: Best practices and benchmarks,'' \emph{arXiv
  preprint arXiv:1903.00954}, 2019.

\bibitem{Arnez2020ACO}
F.~Arnez, H.~Espinoza, A.~Radermacher, and F.~Terrier, ``A comparison of
  uncertainty estimation approaches in deep learning components for autonomous
  vehicle applications,'' \emph{Proc. Workshop Artif. Intell. Safety}, 2020.

\bibitem{kendall_gal}
A.~Kendall and Y.~Gal, ``What uncertainties do we need in {B}ayesian deep
  learning for computer vision?'' in \emph{Int. Conf. Neural Inf. Process.
  Syst. (NeurIPS)}, 2017, pp. 5580--5590.

\bibitem{russell2021multivariate}
R.~L. Russell and C.~Reale, ``Multivariate uncertainty in deep learning,''
  \emph{IEEE Trans. Neural Netw. and Learn. Syst.}, vol.~33, no.~12, pp.
  7937--7943, 2021.

\bibitem{Bishop1994MixtureDN}
C.~M. Bishop, ``Mixture density networks,'' Aston University, Tech. Rep., 1994.

\bibitem{scott1992multivariate}
D.~W. Scott, \emph{Multivariate Density Estimation: Theory, Practice, and
  Visualization}.\hskip 1em plus 0.5em minus 0.4em\relax New York: John Wiley
  \& Sons, 1992.

\bibitem{deconv_density_net}
B.~Chen, M.~Islam, L.~Wang, J.~Gao, and J.~Orchard, ``Deconvolutional density
  network: Free-form conditional density estimation,'' in \emph{36th {AAAI}
  Conf. Artif. Intell.}, 05 2022.

\bibitem{deep_dist_regression}
R.~Li, B.~J. Reich, and H.~D. Bondell, ``Deep distribution regression,''
  \emph{Comput. Statist. \& Data Anal.}, vol. 159, p. 107203, 2021.

\bibitem{tansey2016better}
W.~Tansey, K.~Pichotta, and J.~G. Scott, ``Better conditional density
  estimation for neural networks,'' \emph{arXiv preprint arXiv:1606.02321},
  2016.

\bibitem{Kingma2014}
D.~P. Kingma and M.~Welling, ``Auto-encoding variational {B}ayes,'' in
  \emph{Int. Conf. Learn. Representations ({ICLR})}, 2014.

\bibitem{realnvp}
L.~Dinh, J.~Sohl-Dickstein, and S.~Bengio, ``Density estimation using real
  nvp,'' in \emph{Int. Conf. Learn. Representations (ICLR)}, 2017.

\bibitem{maf}
G.~Papamakarios, T.~Pavlakou, and I.~Murray, ``Masked autoregressive flow for
  density estimation,'' in \emph{Advances Neural Inf. Process. Syst.
  (NeurIPS)}, vol.~30, 2017.

\bibitem{iaf}
D.~P. Kingma, T.~Salimans, R.~Jozefowicz, X.~Chen, I.~Sutskever, and
  M.~Welling, ``Improved variational inference with inverse autoregressive
  flow,'' in \emph{Advances Neural Inf. Process. Syst. (NeurIPS)}, 2016, pp.
  4743--4751.

\bibitem{acharya2022competency}
A.~Acharya, R.~Russell, and N.~R. Ahmed, ``Competency assessment for autonomous
  agents using deep generative models,'' \emph{IEEE/RSJ Int. Conf. Intell.
  Robots and Syst. (IROS)}, pp. 8211--8218, 2022.

\bibitem{acharya2023learning}
A.~Acharya, R.~L. Russell, and N.~Ahmed, ``Learning to forecast aleatoric and
  epistemic uncertainties over long horizon trajectories,'' \emph{IEEE Int.
  Conf. Robot. and Automat. (ICRA)}, pp. 12\,751--12\,757, 2023.

\bibitem{papamakarios2021normalizing}
G.~Papamakarios, E.~Nalisnick, D.~J. Rezende, S.~Mohamed, and
  B.~Lakshminarayanan, ``Normalizing flows for probabilistic modeling and
  inference,'' \emph{J. Machine Learn. Res.}, vol.~22, no.~57, pp. 1--64, 2021.

\bibitem{mackay2003information}
D.~J. MacKay, \emph{Information Theory, Inference, and Learning
  Algorithms}.\hskip 1em plus 0.5em minus 0.4em\relax Cambridge, UK: Cambridge
  University Press, 2003.

\bibitem{beran1977minimum}
R.~Beran, ``Minimum {H}ellinger distance estimates for parametric models,''
  \emph{Ann. Stat.}, vol.~5, no.~3, pp. 445--463, 1977.

\bibitem{sgtd}
P.~Theodossiou, ``Financial data and the skewed generalized t distribution,''
  \emph{Manage. Sci.}, vol.~44, no. 12-part-1, pp. 1650--1661, 1998.

\bibitem{airdrop}
C.~Dever \emph{et~al.}, ``Guided-airdrop vision-based navigation,'' in
  \emph{24th AIAA Aerodynamic Decelerator Syst. Tech. Conf.}, 2017.

\bibitem{splice}
K.~Smith \emph{et~al.}, ``Operational constraint analysis of terrain relative
  navigation for landing applications,'' in \emph{AIAA Scitech 2020 Forum},
  2020.

\bibitem{maggio_ibal}
D.~R. Maggio, C.~Mario, B.~Streetman, T.~Steiner, and L.~Carlone,
  ``Vision-based terrain relative navigation on high altitude balloon and
  sub-orbital rocket,'' in \emph{AIAA SciTech 2023 Forum}, 2023, p. 0875.

\end{thebibliography}

\end{document}